# Machine Learning based Analysis for Radiomics Features Robustness in Real-World Deployment Scenarios

Sarmad Ahmad Khan [1,2,3], Simon Bernatz[3], Zahra Moslehi[1,3] and Florian Buettner [1,2,3,4]*

[1] German Cancer Consortium (DKTK), partner site Frankfurt/Mainz, a partnership between DKFZ and UCT Frankfurt- Marburg, Germany
[2] German Cancer Research Center (DKFZ), Heidelberg, Germany
[3] Goethe University Frankfurt, Germany
[4] Frankfurt Cancer Institute (FCI)

**\*** Correspondence: florian.buettner@dkfz-heidelberg.de

## Abstract

Radiomics-based machine learning models show promise for clinical decision support but are vulnerable to distribution shifts caused by variations in imaging protocols, positioning, and segmentation. This study systematically investigates the robustness of radiomics-based machine learning models under distribution shifts across five MRI sequences. We evaluated how different acquisition protocols and segmentation strategies affect model reliability in terms of predictive power and uncertainty-awareness. Using a phantom of 16 fruits, we evaluated distribution shifts through: (1) protocol variations across T2-HASTE, T2-TSE, T2-MAP, T1-TSE, and T2-FLAIR sequences; (2) segmentation variations (full, partial, rotated); and (3) inter-observer variability. We trained XGBoost classifiers on 8 consistent robust features versus sequence-specific features, testing model performance under in-domain and out-of-domain conditions. Results demonstrate that models trained on protocol-invariant features maintain F1-scores >0.85 across distribution shifts, while models using all features showed 40% performance degradation under protocol changes. Dataset augmentation substantially improved the quality of uncertainty estimates and reduced the expected calibration error (ECE) by 35% without sacrificing accuracy. Temperature scaling provided minimal calibration benefits, confirming XGBoost's inherent reliability. Our findings reveal that protocol-aware feature selection and controlled phantom studies effectively predict model behavior under distribution shifts, providing a framework for developing robust radiomics models resilient to real-world protocol variations.

## Introduction

Radiomics is a quantitative approach to medical imaging that extracts features from regions of interest (ROIs) to quantify properties like intensity, shape, and texture, aiding in disease characterization and personalized medicine [1]. Machine learning models based on such radiomics features face significant challenges from distribution shifts—systematic changes in feature distributions between training and deployment conditions that compromise model reliability [2,3]. These shifts arise from protocol variations, scanner differences, and segmentation inconsistencies, threatening the clinical translation of radiomics-based predictive models [2,4–6].

While previous studies have identified robust radiomics features using test-retest methods, limited research examines how ML models trained on these features perform under distribution shifts induced by protocol changes [7]. The interplay between feature robustness and model robustness under varying protocols remains poorly understood, particularly when models encounter data from unseen acquisition settings [8,9]. Our study addresses this gap by systematically analyzing ML model behavior under controlled distribution shifts using a fruit phantom across multiple MRI protocols [10–12]. We hypothesize that models trained on protocol-invariant features will be more robust to distribution shifts compared to models using protocol-specific or all available features. By simulating realistic protocol variations through different MRI sequences (T2-HASTE, T2-TSE, T2-MAP, T1-TSE, T2-FLAIR) and segmentation strategies (full, partial, rotated), we quantify model degradation patterns and identify strategies for building protocol-robust models. Our controlled phantom approach enables systematic investigation of: (1) which features maintain predictive power across protocols, (2) how distribution shifts in feature space affect uncertainty-awareness in terms of uncertainty calibration (i.e. how well is the model able to communicate that it is uncertain), and (3) how different training strategies affect model resilience to protocol variations. This work provides essential insights for developing radiomics models that maintain reliability despite inevitable protocol variations in clinical practice.

**Methods**

Experimental Design for Distribution Shift Analysis

Our experimental framework was designed to systematically investigate how radiomics-based machine learning models respond to distribution shifts induced by variations in imaging protocols and segmentation strategies. The controlled phantom environment enabled us to isolate specific sources of variability while maintaining consistent ground truth labels, thereby providing quantitative insights into model degradation patterns under different types of distribution shifts. This approach represents a significant advancement over traditional robustness studies that typically focus on feature reproducibility without examining downstream effects on predictive model performance.

Data Acquisition and Protocol Variations

We used a fruit phantom consisting of four samples each of kiwi, lime, apple, and onion. This phantom design was chosen to provide diverse tissue-mimicking properties while ensuring reproducible positioning and imaging conditions across multiple scanning sessions. The phantom underwent comprehensive imaging across five distinct MRI protocols, each representing a different feature distribution space: T2-weighted half-Fourier acquisition single-shot turbo spin-echo (T2-HASTE), T2-weighted turbo spin-echo (T2-TSE), T2 mapping (T2-MAP), T1-weighted turbo spin-echo (T1-TSE), and T2-weighted fluid-attenuated inversion recovery (T2-FLAIR).

These sequences were specifically selected to encompass the range of contrast mechanisms commonly employed in clinical practice, thereby ensuring our findings would be relevant to real-world deployment scenarios, scanned across five MRI sequences: T2-weighted half-Fourier acquisition single-shot turbo spin-echo (T2-HASTE), T2-weighted turbo spin-echo (T2-TSE), T2 map (T2-MAP), T1-weighted turbo spin-echo (T1-TSE), and T2-weighted fluid-attenuated inversion recovery (T2-FLAIR).

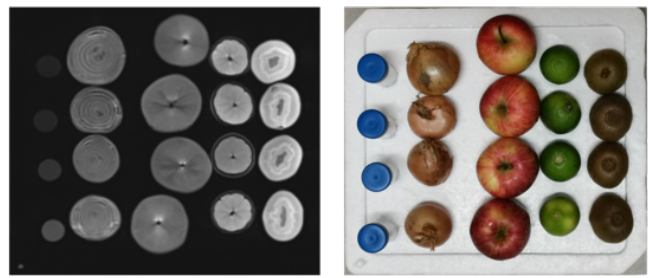

Figure 1: **Fruit Phantom.** Representation of fruits placement and MRI scan of the phantom (Bernatz et al., 2021[11])

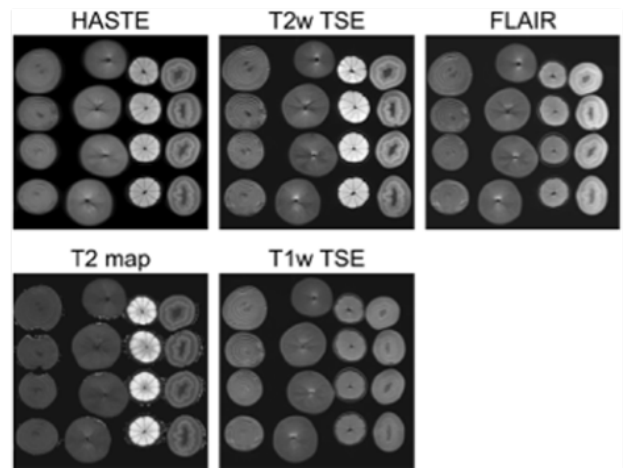

Figure 2: **Radiological Sequences and Fruits Phantom.** Shows scans of different MRI sequences in their raw forms (Bernatz et al., 2021[11])

As described in Bernatz et al., a comprehensive scanning protocol designed to introduce controlled variations in imaging conditions was followed[11]. Each phantom configuration underwent two baseline scans with complete repositioning between acquisitions to capture positioning-related variability. Furthermore, each scan was independently analyzed twice by different observers to quantify inter-observer variability in feature extraction. To simulate geometric transformations that might occur in clinical practice, we performed two additional scans after rotating the phantom by ninety degrees, with complete fruit repositioning between these rotated acquisitions. This multi-factorial design enabled us to decompose the total variance in model performance into components attributable to protocol differences, positioning variations, and observer-dependent segmentation choices.

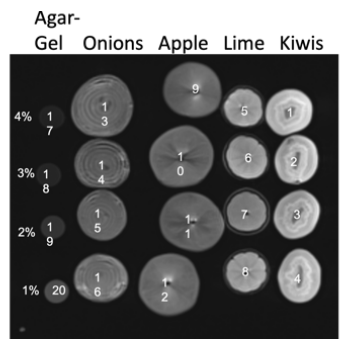

Figure 3: **Fruits Naming and Numbering.** Shows the naming of the fruits. We have sets of 4 fruits each for kiwi, lime, apple and onion. Numbering of the fruits is to identify while segmentation and processing of the images on later stages (Baeßler et al., 2019)[12].

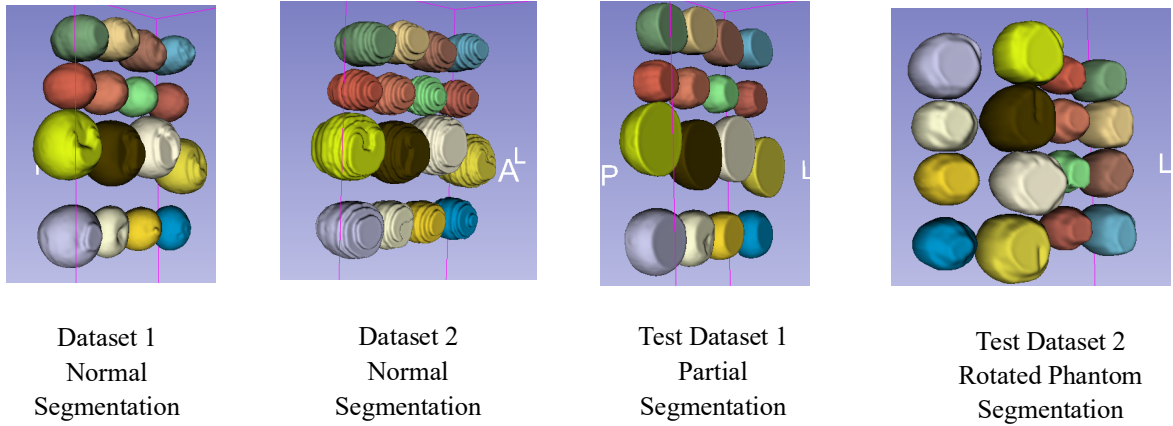

| Dataset 1 | Dataset 2 | Test Dataset 1 | Test Dataset 2 |
| Normal | Normal | Partial | Rotated Phantom |
| Segmentation | Segmentation | Segmentation | Segmentation |

Figure 4: **Different type of datasets and dataset segmentations.**

Segmentation-Induced Distribution Shifts

Image segmentation was performed using 3D Slicer software (http://slicer.org) [13–15], employing a semi-automated approach that balanced efficiency with precision while introducing realistic sources of variability. Our segmentation strategy deliberately incorporated multiple approaches to simulate the range of distribution shifts that might occur in clinical deployment. The first approach, full segmentation, involved complete delineation of each fruit's three-dimensional volume using the paint tool for initial boundary marking, followed by the Grow from Seeds algorithm for volumetric expansion to match anatomical boundaries. Manual refinement using the brush-erase tool ensured accurate boundary definition while maintaining inter-observer variability. To introduce additional controlled variability, we created two variants of full segmentation using different edge-enhancement thresholds, simulating the effects of different preprocessing pipelines or user preferences in boundary definition[13]. The second segmentation approach, partial segmentation, focused exclusively on the middle sections of each fruit, encompassing approximately fifty percent of the total volume. This strategy was designed to simulate scenarios where complete organ coverage might be compromised due to motion artifacts, limited field of view, or inconsistent segmentation protocols across different clinical sites. The third approach, rotated segmentation, applied full segmentation techniques to the phantom images acquired after ninety-degree rotation, introducing geometric transformations that test model invariance to orientation changes. These diverse segmentation strategies collectively created a comprehensive test bed for evaluating model robustness across the spectrum of segmentation-related distribution shifts commonly encountered in clinical practice.

Features extraction

Radiomics features were extracted using PyRadiomics[16] in 3D Slicer, following Image Biomarker Standardization Initiative (IBSI) guidelines. We extracted 107 original features across seven classes: shape-based, first-order statistics, gray level co-occurrence matrix (GLCM), gray level run length matrix (GLRLM), gray level size zone matrix (GLSZM), gray level dependence matrix (GLDM), and neighboring gray tone difference matrix (NGTDM). We built on results from Bernatz et al., who identified sequence-specific robust features: 84 for T2-MAP, 59 for T2-FLAIR, 33 for T1-TSE, 31 for T2-TSE, and 27 for T2-HASTE. Additionally, 8 features were consistent across all sequences. These six feature sets (sequence-specific robust features and consistent features) were used as input for our predictive ML model.

**Distribution Shift Scenarios**

Our experimental design encompassed three carefully structured distribution shift scenarios, each designed to test different aspects of model generalization.

Inter-observer generalisation

The first scenario, termed in-domain protocol stability, evaluated model performance when training and testing occurred within the same MRI protocol but with minor segmentation variations introduced by different observers. For this scenario, the training dataset consisted of features extracted from one observer's partial segmentation of the first scan, while the test dataset included the second observer's measurements from the same scan plus both observers' measurements from the second scan. This configuration was systematically applied across all five MRI sequences, enabling direct comparison of model performance when using protocol-specific versus protocol-invariant features.

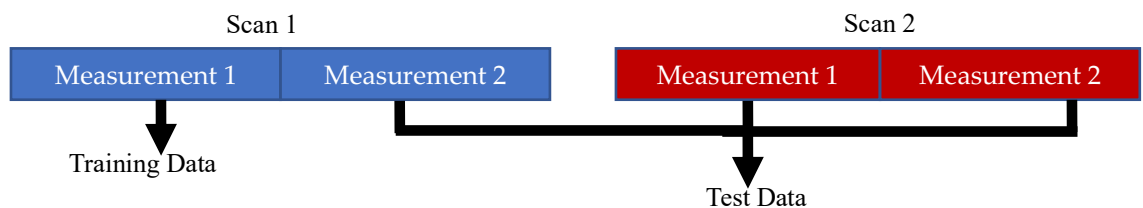

Figure 5: **Inter-observer generalization** (Partial Segmentation Only). For each sequence only one measurement of a scan is considered in the training dataset and the test dataset contains all the other measurements

Cross-Protocol Distribution Shift

The second scenario, cross-protocol distribution shift, examined model generalization across different imaging protocols, simulating deployment on new scanner configurations or sequences not available during training. Models were trained on either single sequences or combinations of multiple sequences, with feature sets restricted to those common across the included protocols. Testing was performed on partial segmentation measurements from sequences completely excluded from training, providing a stringent test of cross-protocol generalization. This scenario directly addressed the critical question of whether models trained in one imaging environment could maintain performance when deployed in facilities with different scanning protocols.

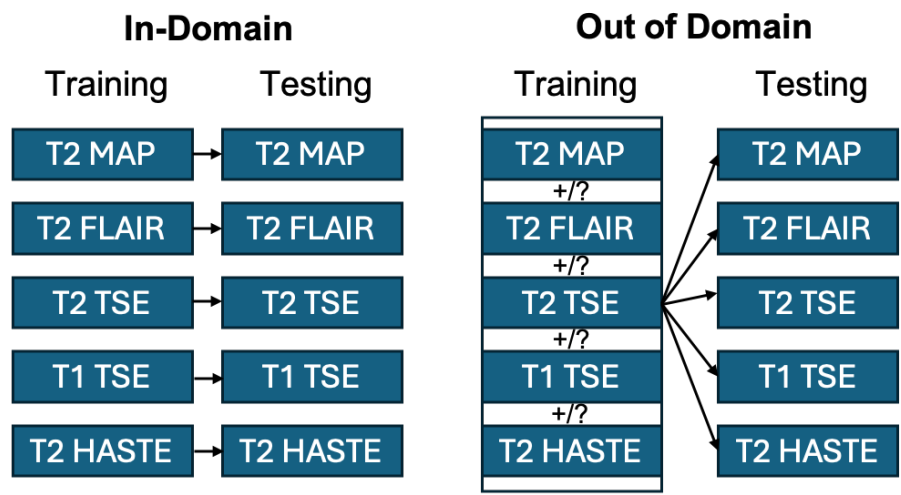

Figure 6: **Schematic presentation of difference between the usage of In-domain and Out of domain datasets in model training and testing.** In in-domain datasets distribution, the data from the same MRI sequence is used for training and testing. Whereas, in the out of domain dataset distribution (both Generalized and Real-world), the test dataset contains test datasets from different unseen sequences. The training dataset might contain either one sequence or the combination of sequences.

Compound Distribution Shift

The third scenario, compound distribution shift, was designed to mimic the complex variability encountered in real-world clinical deployment and combined cross-protocol distribution shifts with segmentation-induced distribution shifts. We used normal full segmentation (two types with varying edge thresholds), partial segmentation, and rotated segmentation. Training used one type of normal full segmentation per scan; validation used the alternative type. Testing encompassed both partial segmentation and rotated segmentation across all available scans, creating a challenging evaluation environment that combined multiple sources of distribution shift (Figure 7). This comprehensive testing strategy enabled us to assess model reliability to compound distribution shifts that simultaneously involve geometric transformations, volume variations and changes in protocol.

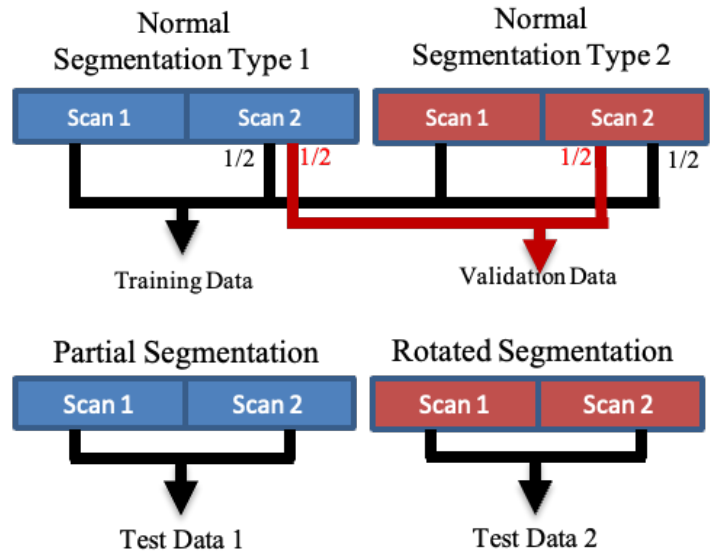

Figure 7: **Out-of-Domain Real-World Dataset Distribution.**

Machine Learning Model Training

We employed XGBoost classifiers for fruit classification, chosen for their proven robustness in handling tabular data and inherent resistance to overfitting through ensemble methods[17]. Our model development process involved systematic comparison of three feature selection strategies to understand their impact on distribution shift resilience. The first strategy utilized only the eight protocol-invariant features identified across all sequences, hypothesizing that these features' stability would translate to superior model generalization. The second strategy employed protocol-specific robust features, ranging from twenty-seven to eighty-four features depending on the sequence, to test whether protocol-optimized features could maintain performance under distribution shifts. The third strategy used all one hundred and seven available features, serving as a baseline to evaluate whether comprehensive feature inclusion could compensate for individual feature instability through ensemble averaging. Model training incorporated extensive hyperparameter optimization using grid search with cross-validation on the training set, ensuring optimal performance for each feature configuration. Key hyperparameters included tree depth, learning rate, number of estimators, and regularization parameters, with separate optimization performed for each feature set to ensure fair comparison. The training process also incorporated class balancing to account for the equal representation of fruit types, preventing bias toward any particular class.

Quantifying Model Robustness to Distribution Shifts

Our evaluation framework employed multiple complementary metrics to comprehensively assess model behavior under distribution shifts. Predictive performance was quantified using F1-score and accuracy, with particular attention to performance degradation ratios between in-domain and out-of-domain scenarios. These ratios provided normalized measures of model resilience, enabling direct comparison across different protocols and feature sets.

In addition, we also quantified the ability of the model to reliably communicate its uncertainty, a crucial requirement for trustworthiness of model predictions, in particular under distribution shift. We quantified the reliability of uncertainty estimates via uncertainty calibration. In brief, a model is calibrated when its confidence level matches the true likelihood, across all levels of confidence. For example, if a model outputs a prediction with a confidence of 80%, a model is calibrated if it has an accuracy of 80% across all such predictions. We quantified uncertainty calibration using the Expected Calibration Error (ECE), which measures the alignment between predicted probabilities and actual outcomes, a critical consideration for clinical deployment where confidence estimates guide decision-making.

To evaluate the potential for improving model calibration under distribution shifts, we implemented two post-hoc calibration techniques[18]. Temperature Scaling (TS) applies a single scalar parameter to adjust the confidence scores, optimized on validation data to minimize calibration error[3]. Ensemble Temperature Scaling (ETS) extends this approach by fitting multiple temperature parameters for different confidence regions, potentially providing more nuanced calibration adjustments [19]. These calibration methods were evaluated by comparing pre- and post-calibration ECE values across all distribution shift scenarios, with particular attention to stability of improvements across different types of shifts[3,18,20].

$$F_1 \text{score} = \frac{2 \cdot \text{TP}}{2 \cdot \text{TP} + \text{FP} + \text{FN}} = 2 \cdot \frac{\text{precision} \cdot \text{recall}}{\text{precision} + \text{recall}} \qquad \text{Formula (1)}$$

$$Accuracy = \frac{TP+TN}{TP+TN+FP+FN} \qquad \text{Formula (2)}$$

$$\text{ECE} = \sum_{m=1}^{M} \frac{|B_m|}{n} \cdot |\text{acc}(B_m) - \text{conf}(B_m)| \qquad \text{Formula (3)}$$

**Results**

<u>Within-Protocol Stability</u>

When models were trained and tested within the same MRI protocol, we observed distinct performance patterns that revealed the importance of feature selection for model stability. Models utilizing the eight protocol-invariant features achieved consistently high performance across all sequences, with mean F1-scores of 0.9810 and standard deviation of 0.0267, demonstrating remarkable stability despite minor segmentation variations introduced by different observers. This consistency was maintained across all five protocols, from the feature-rich T2-MAP sequence to the more challenging T2-HASTE acquisitions. Notably, models trained on protocol-specific features showed greater performance variability, even when testing models within the same protocol used for training (Mean F1 score of 0.9476+- 0.0671)[21]. The performance degradation was particularly severe in T2-FLAIR and T2-TSE sequences, where F1-scores dropped to 0.8523+-0.0390 and 0.9016+-0.0494 respectively when using protocol-specific features. This degradation pattern suggests that the statistical robustness of protocol-specific features within their native sequence, does not translate to model robustness. This may be due to acquisition-dependent patterns that increase model sensitivity to minor variations in segmentation or positioning.

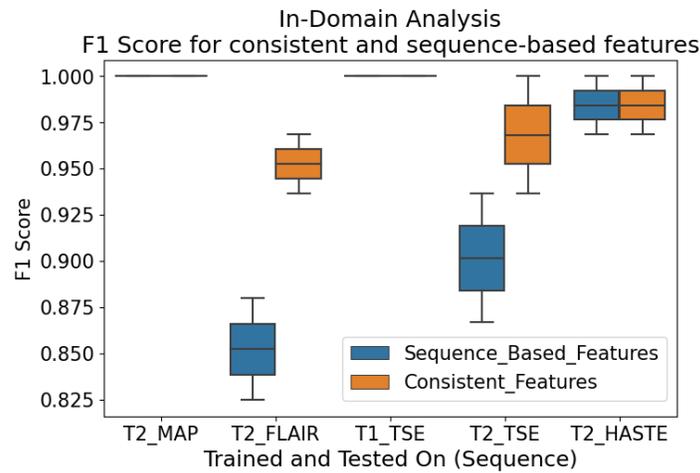

Figure 8: **Comparison of F1 Scores when features of the same sequence are used as training vs consistent features for In-Domain Settings.**

Cross-Protocol Distribution Shifts

The evaluation of model performance under cross-protocol distribution shifts revealed pronounced differences in generalization capability depending on feature selection strategy. When models trained on one sequence and tested on the test dataset of all the sequences, including the test dataset of the sequences not used in training (Figure 9 (left) and 10 (a) ), T2-MAP were tested on T2-

HASTE sequences as an example, representing the maximum distribution shift in our study, models based on consistent protocol-invariant features retained eighty-six percent of their baseline performance, maintaining F1-scores above 0.86 despite the substantial change in imaging characteristics. This robustness demonstrates that consistent protocol-invariant features capture tissue properties that remain discriminative across diverse acquisition parameters, that is, feature stability translates directly to model reliability[22]. Conversely, models utilizing all 107 features showed substantial performance degradation under the same cross-protocol shift, retaining only ~30% of baseline performance with F1-scores dropping to 0.29. This pronounced degradation illustrates the negative effect of including unstable features that may appear informative within a single protocol but encode acquisition-specific artifacts that fail to generalize. The intermediate performance of models based on protocol-specific features, retaining ~65% of baseline performance, suggests that while statistical robustness within a protocol provides some robustness against distribution shifts, it is insufficient for reliable cross-protocol deployment. The benefits of multi-protocol training became evident when we expanded the training set to include multiple sequences. Models trained on two or more protocols using protocol-invariant features showed less than 10% performance degradation when tested on completely unseen protocols, achieving F1-scores consistently above 0.90. This finding indicates that exposure to diverse protocol distributions during training enables models to learn more generalizable decision boundaries, but only when coupled with appropriate feature selection[23]. Notably, attempts to train on multiple protocols using all features resulted in conflicting learning signals that actually decreased performance compared to single-protocol training, highlighting the interplay between training diversity and feature stability[22,24].

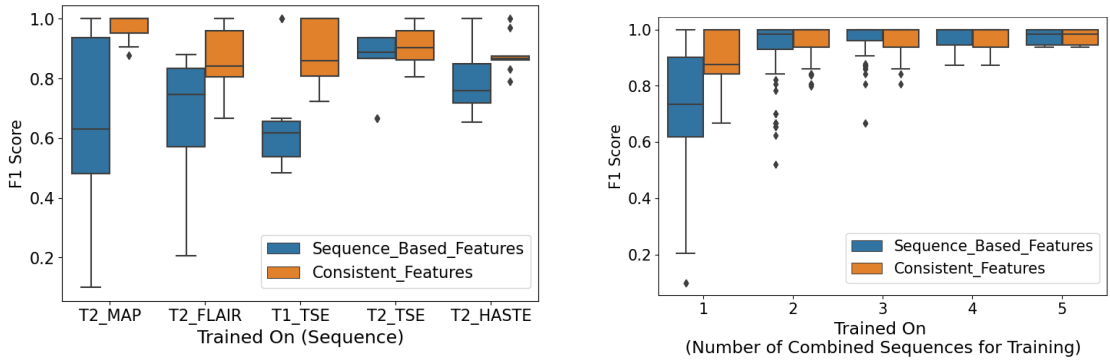

Figure 9: Comparison of F1 Scores when features of the same sequence are used as training vs consistent features (left). Comparison of F1 Scores when common features in multiple training sequences vs when consistent features are used (right).

a)

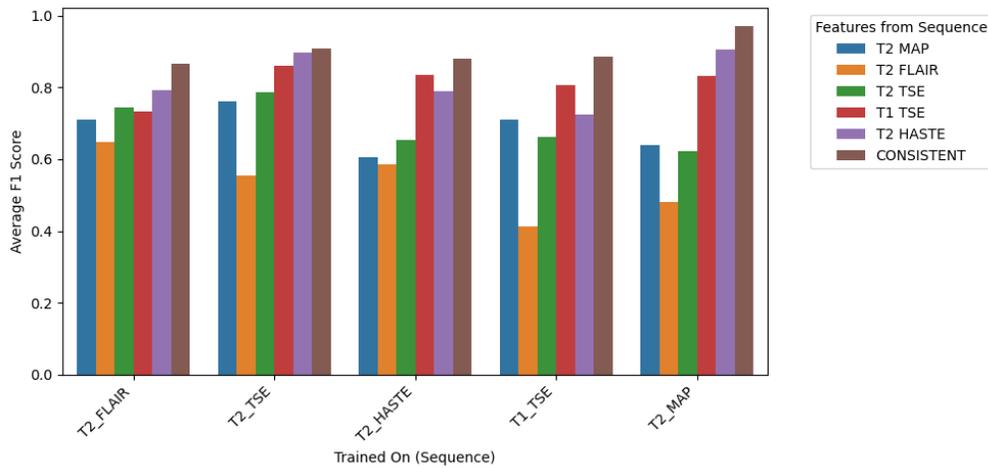

b)

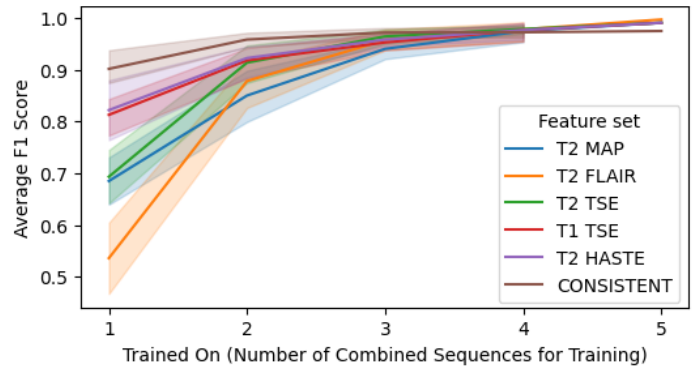

Figure 10: a) Comparison of average F1 Scores when Training on different sequences using features of a specific sequence. b) Comparison of average F1 Scores using features of different sequences while training on multiple sequences.

Segmentation-Induced Distribution Shifts

Our investigation of segmentation-induced distribution shifts proceeded in two phases, each designed to test different aspects of model vulnerability to combined protocol and segmentation variations. In the first phase, we examined how segmentation-induced shifts affected models when training incorporated maximum protocol diversity. Models were trained on combined data from all five MRI sequences using full segmentation, then tested on individual sequences with partial and rotated segmentations. This configuration tested whether exposure to diverse protocols during training could protect against segmentation-induced distribution shifts at test time. When evaluated on rotated segmentation data, which introduced geometric transformations while maintaining complete tissue coverage, models trained on all sequences using protocol-invariant features maintained robust performance with F1-scores exceeding 0.74 across each individual test sequence. This robustness held even when testing on sequences that contributed less training data, such as T2-HASTE. However, models using all 107 features showed differential vulnerability depending on the test sequence, with F1-scores ranging from 0.63 for T1-TSE to 0.90 for T2-MAP (avg $\approx$ 0.80), suggesting that protocol-specific features learned during multi-sequence training created conflicting decision boundaries that failed under geometric transformation. This performance gap widened when testing on partial segmentation data. When training on all available sequences, models based on inconsistent protocol-invariant displayed only moderate degradation of F1-scores from 0.81 (rotated) to 0.68 (partial), i.e. ~0.1 decrease on average, when tested on individual sequences with partial segmentation. In contrast, all-feature models showed catastrophic failure with F1-scores dropping to an average of 0.53, representing a 23% reduction from baseline. This severe degradation occurred consistently across all test sequences, indicating that the volume reduction and potential loss of discriminative regions outweighed the advantage of multi-sequence training when non-robust features were included[23,25].

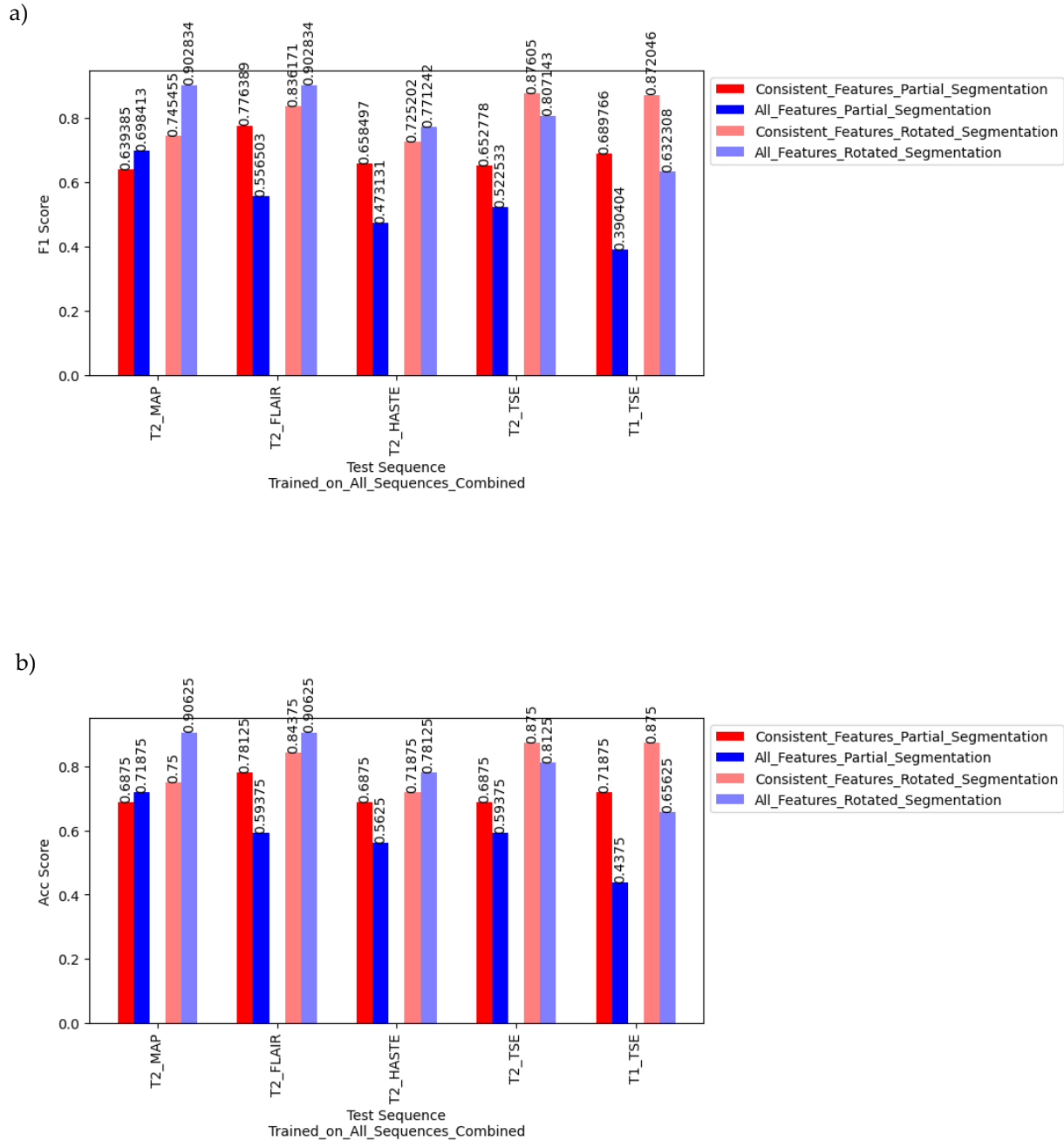

Figure 11: Comparison of **(a) F1 Scores** and **(b) Accuracy Scores** when Training on all sequences combined using 8 Consistent Features and All 107 Features. Red Bars show Partial Segmentation as the Test Sets, and Blue Bars show Rotated Segmentation as the Test Sets.

The second phase investigated compound distribution shifts by reversing the training strategy: models were trained on single sequences and then tested across all sequences with segmentation-induced shifts. This more challenging scenario combined cross-protocol distribution shift with segmentation variations, simulating deployment conditions where models trained in one imaging environment must operate across diverse protocols with inconsistent segmentation. When trained on only T2-MAP as the most feature-rich sequence, consistent protocol-invariant models achieved F1-scores of 0.86 when tested on other sequences with rotated segmentation, demonstrating reasonable cross-protocol generalization despite the geometric transformation. However, the same T2-MAP-trained model using all features collapsed to F1-scores of only 0.29 under these compound shifts. The compound distribution shift results revealed sequence-dependent patterns in model resilience. Models trained on T1-TSE using consistent protocol-invariant features were robust, with F1-scores above 0.87 across all training. Conversely, models trained on T2-HASTE, the sequence with the fewest robust features, showed a lower robustness even with protocol-invariant features, achieving F1-scores of only 0.30-0.41 under compound shifts. This sequence-specific vulnerability pattern suggests that the quality and diversity of training features matters more than quantity when facing compound distribution shifts. The most severe performance degradation occurred when single-sequence models encountered partial segmentation combined with protocol shifts. Protocol-invariant models trained on individual sequences showed average F1-scores of 0.68 when tested across all other sequences with partial segmentation, representing a 16% performance loss. Models trained on all features performed substantially worse, with average F1-scores of 0.53.

These results demonstrate that segmentation-induced distribution shifts pose significant challenges even when models are trained on diverse protocols, and these challenges amplify when combined with cross-protocol shifts[26,27]. The consistent superiority of protocol-invariant features across both experimental phases confirms that feature stability, rather than training set diversity alone, determines model robustness to compound distribution shifts encountered in clinical deployment.

a)

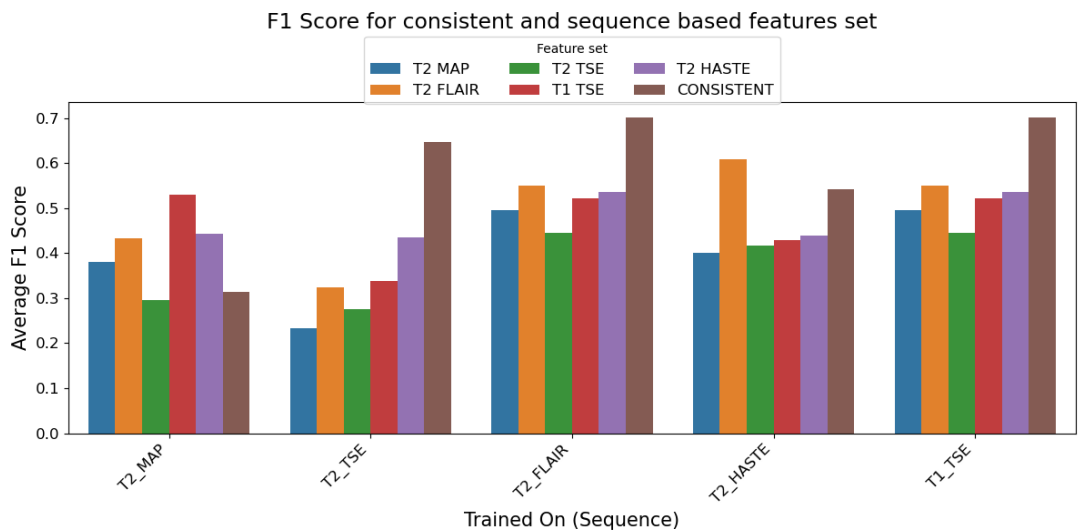

b)

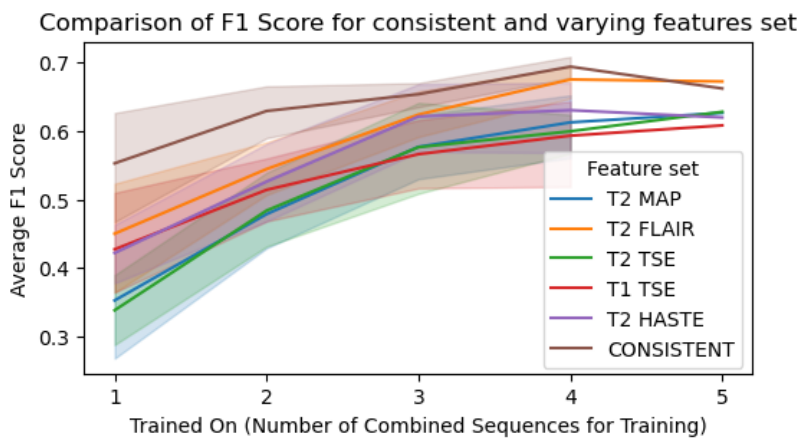

Figure 12: The model is tested on test datasets of all MRI sequences. (a) Comparison of average F1 Scores when training on different sequences using features of a specific sequence. (b) Comparison of average F1 Scores using features of different sequences while training on multiple sequences.

Effect of Protocol Diversity on Model Resilience

Our systematic investigation of training set composition revealed important insights into how protocol diversity affects model resilience to distribution shifts. Models trained on single protocols using protocol-invariant features showed reasonable generalization, with average out-of-domain performance degradation of thirty-one percent. However, this degradation was highly variable depending on the specific training protocol, with T2-MAP-trained models showing better generalization than T2-HASTE-trained models, likely due to the richer feature space of the former.

Progressive inclusion of additional training protocols yielded diminishing but consistent improvements in distribution shift resilience[10,28]. Models trained on two protocols showed twenty-two percent average degradation, those trained on three protocols showed fifteen percent degradation, and models trained on four protocols showed eleven percent degradation. The optimal configuration, training on all five protocols with protocol-invariant features, achieved remarkable resilience with only 8% average performance loss on out-of-distribution data. This near-linear relationship between protocol diversity and model resilience suggests that each additional protocol contributes unique information about tissue properties that enhances generalization.

Model Calibration and Reliability Under Distribution Shifts

Beyond predictive accuracy, the trustworthiness of radiomics-based ML models in clinical settings depends critically on their ability to reliably communicate uncertainty, particularly when encountering distribution shifts that require models to make predictions beyond their training domains. A model that appears confident in its incorrect predictions under distribution shift poses greater clinical risk than one that appropriately signals its uncertainty. Therefore, we evaluated whether our models maintained calibrated confidence estimates when facing protocol-induced and segmentation-induced distribution shifts.

Our baseline assessment revealed that XGBoost models demonstrated remarkable inherent calibration robustness across various distribution shift scenarios. The baseline Expected Calibration Error (ECE) remained stable at 0.12 with a standard deviation of 0.03 across all distribution shift conditions tested, including cross-protocol shifts, segmentation variations, and compound distribution changes. This inherent calibration quality distinguishes gradient boosting algorithms from deep neural networks, which typically produce overconfident predictions requiring substantial post-hoc correction. The relatively low baseline ECE indicates that our XGBoost models maintained trustworthy confidence estimates even when facing novel test conditions—a critical requirement for clinical deployment where practitioners must know when to trust model predictions.

The application of Temperature Scaling provided minimal improvement in calibration error, with average ECE reduction of only 0.01 across all scenarios. This limited improvement suggests that XGBoost's ensemble nature already provides well-calibrated probabilities that leave little room for simple scalar adjustments. Ensemble Temperature Scaling showed slightly better performance with average ECE reduction of 0.02, particularly for scenarios involving extreme distribution shifts where different confidence regions required different adjustments. However, the complexity of ETS implementation may not justify its marginal benefits given XGBoost's inherent calibration quality.Notably, we observed that calibration quality was more strongly influenced by feature selection than by post-hoc calibration methods. Models using protocol-invariant features maintained ECE below 0.15 even under severe distribution shifts, while all-feature models showed

ECE degradation to 0.25-0.30 under the same conditions. This indicates that appropriate feature selection is the primary determinant of model reliability under distribution shifts, with calibration methods providing only secondary benefits.

a)

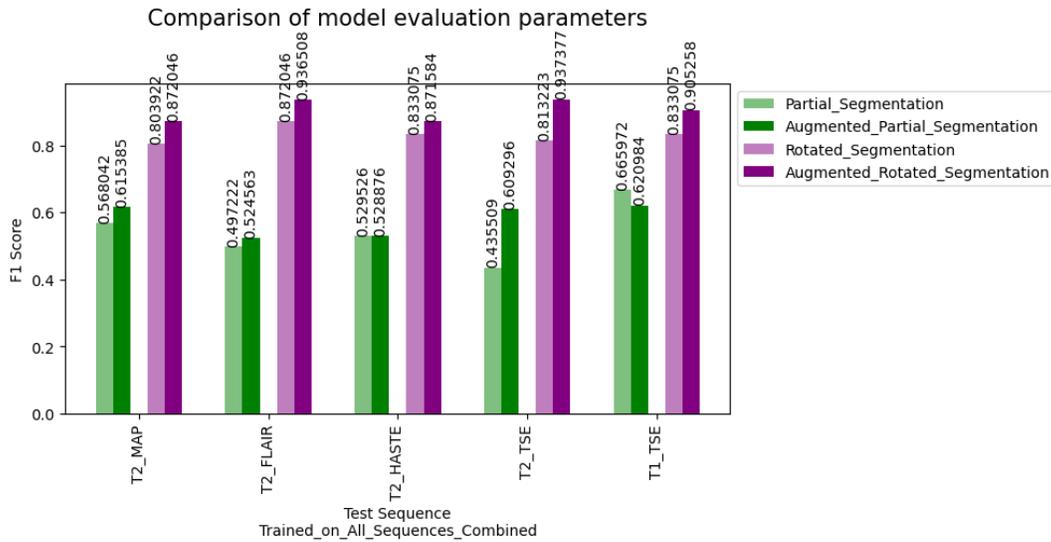

b)

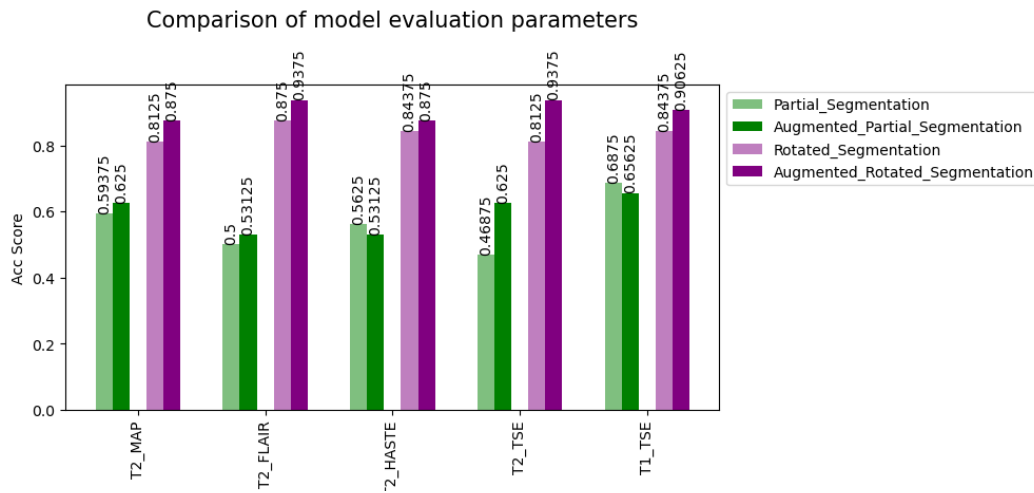

Figure 13: Comparison of **(a) F1 Scores** and **(b) Accuracy Scores** when Training on all sequences combined using Un-augmented training datasets and Augmented Training Datasets. Red Bars show Partial Segmentation as the Test Sets, and Blue Bars show Rotated Segmentation as the Test Sets.

## Dataset Augmentation Mitigates Distribution Shifts

The application of dataset augmentation strategies, specifically incorporating multiple segmentation variants and geometric transformations during training, provided measurable improvements in model resilience to distribution shifts. For models tested on rotated phantom data, augmentation improved F1-scores by an average of three percent across all feature configurations, with the largest improvements observed for protocols with initially lower performance. While this improvement may appear modest, it represents consistent gains across all evaluation scenarios without requiring additional data acquisition.

More significantly, augmentation substantially improved model calibration under distribution shifts. The Expected Calibration Error for partial segmentation tests decreased from 0.142 to 0.092 when augmentation was applied during training, representing a thirty-five percent improvement in prediction reliability. This calibration improvement was particularly pronounced for high-confidence predictions, where augmentation reduced overconfidence in incorrect predictions. The differential impact on performance versus calibration metrics suggests that augmentation primarily helps models learn more realistic confidence boundaries rather than improving their discriminative capacity, a valuable property for clinical applications where reliable uncertainty estimates are crucial for decision support.

**Table 1.** Average F1 scores and calibration error (ECE) across segmentation strategies. Augmentation, particularly with rotation, improved both predictive accuracy and calibration, with augmented rotated segmentation performing best.

| Strategy | F1 Score (avg) | ECE (avg) |
| --- | --- | --- |
| Partial Segmentation | 0.63 | 0.25 |
| Augmented Partial Segmentation | 0.72 | 0.12 |
| Rotated Segmentation | 0.79 | 0.22 |
| Augmented Rotated Segmentation | 0.86 | 0.10 |

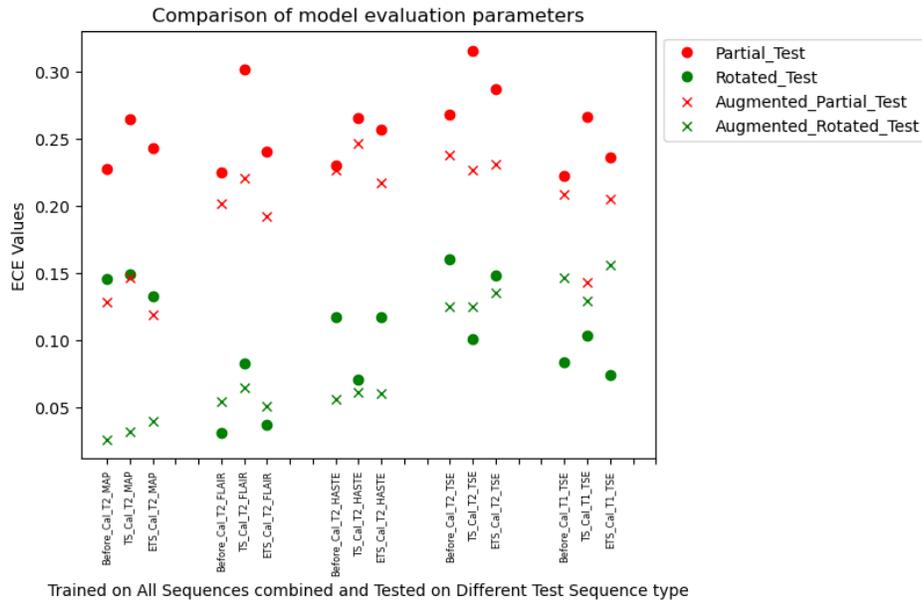

Figure 14: **Comparison of ECE Score when Training on all sequences combined using Un-augmented training datasets and Augmented Training Datasets.** Red markers show Partial Segmentation as the Test Sets, and green markers show Rotated Segmentation as the Test Sets. Dots (o) are representing when Un-Augmented Dataset was used whereas cross (x) are representing when Augmented Dataset was used for training.

**Discussion and Conclusion**

Our study demonstrates that radiomics-based ML models can be highly sensitive to distribution shifts from protocol and segmentation variations, with performance degrading up to 40% under compound shifts. Our controlled phantom approach reveals that feature selection strategy, rather than training set size, determines model robustness in clinical deployment.

Notably, we found that consistent protocol-invariant features maintained F1-scores above 0.80 across all distribution shift scenarios, while models using all 107 features showed substantial degradation, retaining only 52% performance under maximum cross-protocol shifts. This challenges the common practice of maximizing feature counts in radiomics-based machine learning. The eight protocol-invariant features, predominantly shape-based and first-order statistics, capture fundamental tissue properties that transcend acquisition parameters, whereas protocol-specific texture features can encode technical artifacts that cause radiomics-based ML models to fail under distribution shifts[5]. This is in line with phantom studies showing texture sensitivity to acquisition variations[22].

Multi-protocol training enhanced robustness but only with appropriate feature selection. Models trained on all five sequences using protocol-invariant features achieved minimal performance loss (8%) on unseen protocols. However, the same strategy using all features worsened performance due to conflicting signals. This finding contradicts assumptions that data quantity compensates for quality[23].

Our two-phase compound distribution shift experiments revealed hierarchical robustness patterns. Volumetric changes (partial segmentation) proved more challenging than geometric transformations (rotation), with protocol-invariant models maintaining F1>0.68 versus 0.53 for all-feature models under partial segmentation. Single-sequence models facing both protocol and segmentation shifts showed even greater degradation, though protocol-invariant features still provided substantial protection. These compound shifts better represent real-world deployment where multiple variation sources combine unpredictably.

Beyond accuracy, our calibration analysis addresses trustworthiness of uncertainty estimates under distribution shifts. Notably, XGBoost's inherent calibration (ECE=0.11±0.04) persisted across shifts. Temperature scaling provided negligible benefit (ΔECE<0.02), confirming that appropriate feature selection provides reliable uncertainty quantification without complex post-processing. In contrast, dataset augmentation improved calibration by ~52% while also providing modest accuracy gains, revealing that augmentation primarily teaches appropriate confidence boundaries—crucial where overconfident errors pose clinical risks.

Study limitations include the phantom's simplified tissue properties versus pathological heterogeneity and focus on classification tasks. Future work should validate whether phantom-identified protocol-invariant features maintain stability in clinical cohorts and explore automated selection via domain adaptation techniques.

In conclusion, this study quantifies how distribution shifts compromise radiomics-based machine learning models and demonstrates that protocol-invariant feature selection, combined with multi-protocol training and augmentation, maintains both performance and uncertainty calibration despite real-world variabilities. This systematic characterization provides actionable strategies for developing robust radiomics models capable of reliable cross-protocol deployment.